\documentclass[runningheads]{llncs}
\usepackage{graphicx}
\usepackage{cite}
\usepackage{amsmath,amssymb,amsfonts}
\usepackage{algorithmic}
\usepackage{graphicx}
\usepackage{textcomp}
\usepackage{adjustbox}
\usepackage{multirow}
\usepackage{bbold}
\usepackage{siunitx}
\usepackage{hyperref}       
\usepackage{xcolor}
\usepackage{sidecap}

\usepackage{multirow}
\usepackage[normalem]{ulem}

\usepackage{booktabs}
\usepackage[mathscr]{eucal}

\hypersetup{
    colorlinks,
    linkcolor={blue!50!black},
    citecolor={blue!50!black},
    urlcolor={blue!50!black}
}
\usepackage{amsmath, bm}
\usepackage{hyperref}

\usepackage{pifont}

\usepackage[misc]{ifsym}

\usepackage{floatrow}
\newfloatcommand{capbtabbox}{table}[][\FBwidth]

\begin{document}

\title{PitVQA: Image-grounded Text Embedding LLM for Visual Question Answering in Pituitary Surgery}
\titlerunning{PitVQA: LLM for Visual Question Answering in Pituitary Surgery}

\author{Runlong He\inst{1,2}\textsuperscript{(\Letter)} \and
Mengya Xu\inst{3} \and
Adrito Das\inst{1} \and
Danyal Z. Khan\inst{1,4} \and
Sophia Bano\inst{1,5} \and
Hani J. Marcus\inst{1,4} \and
Danail Stoyanov\inst{1,5} \and
Matthew J. Clarkson\inst{1,2} \and
Mobarakol Islam\inst{1,2}\textsuperscript{(\Letter)}
}

\authorrunning{R He et al.}
%

\institute{Wellcome/EPSRC Centre for Interventional and Surgical Sciences (WEISS), University College London, UK \and 
Dept of Medical Physics \& Biomedical Engineering, University College London, UK \and
Dept of Electronic Engineering, The Chinese University of Hong Kong, Hong Kong \and
Dept of Neurosurgery, National Hospital for Neurology and Neurosurgery, UK \and
Dept of Computer Science, University College London, UK \\
\email{runlong.he.23@ucl.ac.uk, mobarakol.islam@ucl.ac.uk}}

\maketitle              
\begin{abstract}

Visual Question Answering (VQA) within the surgical domain, utilizing Large Language Models (LLMs), offers a distinct opportunity to improve intra-operative decision-making and facilitate intuitive surgeon-AI interaction. However, the development of LLMs for surgical VQA is hindered by the scarcity of diverse and extensive datasets with complex reasoning tasks. Moreover, contextual fusion of the image and text modalities remains an open research challenge due to the inherent differences between these two types of information and the complexity involved in aligning them. This paper introduces PitVQA, a novel dataset specifically designed for VQA in endonasal pituitary surgery and PitVQA-Net, an adaptation of the GPT2 with a novel image-grounded text embedding for surgical VQA. PitVQA comprises 25 procedural videos and a rich collection of question-answer pairs spanning crucial surgical aspects such as phase and step recognition, context understanding, tool detection and localization, and tool-tissue interactions. PitVQA-Net consists of a novel image-grounded text embedding that projects image and text features into a shared embedding space and GPT2 Backbone with an excitation block classification head to generate contextually relevant answers within the complex domain of endonasal pituitary surgery. Our image-grounded text embedding leverages joint embedding, cross-attention and contextual representation to understand the contextual relationship between questions and surgical images. We demonstrate the effectiveness of PitVQA-Net on both the PitVQA and the publicly available EndoVis18-VQA dataset, achieving improvements in balanced accuracy of 8\% and 9\% over the most recent baselines, respectively. Our code and dataset is available at~\url{https://github.com/mobarakol/PitVQA}. 

\end{abstract}

\section{Introduction}

Pituitary surgery, particularly through the endonasal approach, is a complex and delicate procedure that demands high precision and situational awareness~\cite{khan2023current}. Surgeons must navigate through critical anatomical structures, requiring not only an in-depth understanding of the surgical field but also the ability to adapt to dynamic intra-operative conditions~\cite{das2023multi}. In this context, the application of Visual Question Answering (VQA) technologies can offer substantial benefits, such as providing instant information on surgical phases and steps, tool usage, and tissue interactions, as well as offering predictive guidance on forthcoming phases, steps, and instrument requirements, thus enhancing the surgical workflow. The integration of Artificial Intelligence (AI), specifically large language model (LLM) or vision language models (VLM), into the operating room, promises to revolutionize surgical practices by providing real-time decision support, enhancing surgical precision, and fostering a more intuitive interaction between surgeons and technology~\cite{lawson2023artificial,decker2023large,seenivasan2023surgicalgpt}. A pivotal aspect of this integration is the development of systems capable of understanding and responding to complex visual and procedural contexts akin to human experts. VQA emerges as a promising field in this regard, especially within the surgical domain, where it can significantly augment intra-operative decision-making processes and post-operative surgical education~\cite{antol2015vqa,hudson2019gqa,seenivasan2022surgical, seenivasan2023surgicalgpt,bai2023cat}.

The question-answer pairs in the VQA dataset include questions related to surgical images and correct answers. The surgical VQA dataset is crucial for building large language models (LLM) or vision language models (VLM) that can understand and reason about surgical images, answer questions related to surgical procedures and assist surgeons in performing complex tasks in the operating room. Several existing VQA datasets, such as EndoVis18-VQA~\cite{seenivasan2022surgical}, Cholec80-VQA~\cite{seenivasan2022surgical}, SSG-VQA~\cite{yuan2024advancing}, focus on nephrectomy and laparoscopic cholecystectomy surgeries. These datasets contain questions such as identification tasks (e.g., ``What is the name of the white object that is retracted by the top-mid instrument?''~\cite{yuan2024advancing}), and phase recognition (e.g., ``What is the surgical phase of the image?''~\cite{seenivasan2023surgicalgpt}). However, most of these datasets are limited in their representation of complex surgical tasks, size, and diversity. There are also a couple of VQA models associated with these datasets, including  SurgicalVQA~\cite{seenivasan2022surgical}, SurgicalGPT~\cite{seenivasan2023surgicalgpt}, Surgical-VQLA~\cite{bai2023surgicalvqla}, SSG-VQA-Net~\cite{yuan2024advancing}. Nonetheless, these models often utilize poor image-text fusion and naive classification heads to convert the language generation model into VQA classification.

In this paper, we introduce a large and diverse surgical VQA dataset and design a VQA LLM network for pituitary surgery. Our contribution can be summarized as below:

\begin{enumerate}
    \item[--] Builds a PitVQA dataset, a specialized dataset focused on VQA in the context of endonasal pituitary surgery, featuring 25 procedural videos and extensive Q\&A pairs addressing key surgical concepts.
    
    \item[--] Develops a PitVQA-Net, an adaptation of GPT2~\cite{radford2019language} incorporating a novel image-text embedding and gated-attention excitation block (EB) classification head for surgical VQA.
    
    \item[--] Design a image-grounded text embedding approach within PitVQA-Net, enhancing contextual alignment between surgical questions and imagery.
    
    \item[--] Validates PitVQA-Net's superior surgical VQA performance on both PitVQA and EndoVis18-VQA datasets, proving its efficacy in supporting surgeon-AI interactions and decision-making.
\end{enumerate}

\section{Method}

\subsection{Preliminaries}

Vision-language processing seeks to establish a deep understanding of the connection between visual content and natural language. Contrastive Language-Image Pre-training (CLIP)~\cite{radford2021learning} and Bootstrapping Language-Image Pre-training (BLIP)~\cite{li2022blip} are two most significant vision-language processing models that have revolutionized this field. CLIP excels in learning robust image and text representations from large, uncurated datasets, enabling tasks like zero-shot image classification and image-text similarity search. BLIP takes this further by introducing a bootstrapping approach where captions are automatically generated and then filtered. In this work, we utilize BLIP to design our Image-grounded Text Embedding for the surgical VQA task. On the other hand, there is evidence that the squeeze \& excitation (SE)~\cite{hu2018squeeze} block, a form of lightweight gated attention mechanism, enhances the representation power of the network and boosts the prediction accuracy~\cite{roy2018concurrent}. We have designed an excitation block (EB) using gated attention and integrated it into the classification head to amplify the significant features in the model learning.

\subsection{Proposed Method: PitVQA}
\subsubsection{PitVQA Dataset:}
\label{pitvqa_dataset}

\begin{figure}[t]
    \centering
    \includegraphics[width=1\textwidth]{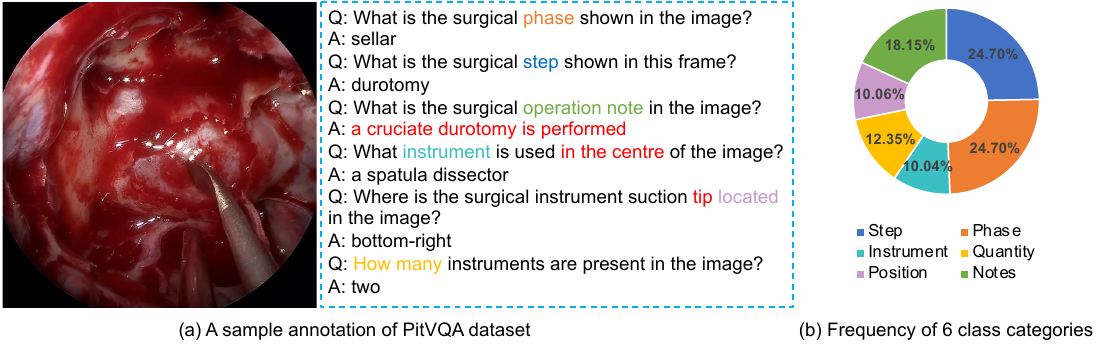}
    \caption{PitVQA dataset of visual questions answering for pituitary surgery. There are overall 59 classes in the 6 class categories of phases, steps, instruments, quantity, positions and operation notes.}
    \label{fig:pitvqa_dataset}
\end{figure}

Our PitVQA dataset comprises 25 videos of endoscopic pituitary surgeries from the The National Hospital of Neurology and Neurosurgery in London, United Kingdom. All patients provided informed consent, and the study was registered with the local governance committee. The surgeries were recorded using a high-definition endoscope (Karl Storz Endoscopy) with a resolution of 720p and stored as MP4 files. All videos were annotated for the surgical phases, steps, instruments present and operation notes guided by a standardised annotation framework, which was derived from a preceding international consensus study on pituitary surgery workflow~\cite{marcus2021pituitary}. Annotation was performed collaboratively by 2 neurosurgical residents with operative pituitary experience and checked by an attending neurosurgeon. We extracted image frames from each video at 1 fps and removed any frames that were blurred or occluded. Ultimately, we obtained a total of 109,173 frames, with the videos of minimum and maximum length yielding 2,443 and 7,179 frames, respectively. We acquired frame-wise question-answer pairs for all the categories of the annotation. Overall, there are 884,242 question-answer pairs from 109,173 frames, which is around 8 pairs for each frame. There are 59 classes overall, including 4 phases, 15 steps, 18 instruments, 3 variations of instruments present in a frame, 5 positions of the instruments, and 14 operation notes in the annotation classes. The length of the questions ranges from a minimum of 7 words to a maximum of 12 words.
A comparison of the unique classes  between our PitVQA and a publicly available dataset of
EndoVis18-VQA is presented in the Table~\ref{table:daraset_comparison}. A sample frame and corresponding Q\&A pairs are presented in Fig.~\ref{fig:pitvqa_dataset}(a). The class frequency distribution is illustrated in Fig.~\ref{fig:pitvqa_dataset}(b), where the lowest and highest classes are instruments and phases with 10.04\% and 24.70\%.

\begin{table}[t]
\centering
\caption{The comparison between our PitVQA and a publicly available dataset of EndoVis18-VQA~\cite{seenivasan2023surgicalgpt}.}
\label{table:daraset_comparison}
\begin{tabular}{clclllcl}
\hline
\multicolumn{2}{c}{Dataset}               & \multicolumn{2}{c}{EndoVis18-VQA} &  &  & \multicolumn{2}{c}{PitVQA} \\ \hline
\multicolumn{2}{c}{Average length (words)}        & \multicolumn{2}{c}{5.8}            &  &  & \multicolumn{2}{c}{10.3}    \\
\multicolumn{2}{c}{Average \#Questions}   & \multicolumn{2}{c}{5.0}            &  &  & \multicolumn{2}{c}{8.1}     \\ \hline
\multicolumn{2}{c}{\#Steps}               & \multicolumn{2}{c}{0}              &  &  & \multicolumn{2}{c}{15}      \\
\multicolumn{2}{c}{\#Phases}              & \multicolumn{2}{c}{0}              &  &  & \multicolumn{2}{c}{4}       \\
\multicolumn{2}{c}{\#Instruments/Objects} & \multicolumn{2}{c}{1}              &  &  & \multicolumn{2}{c}{18}      \\
\multicolumn{2}{c}{\#Quantity}            & \multicolumn{2}{c}{0}              &  &  & \multicolumn{2}{c}{3}       \\
\multicolumn{2}{c}{\#Positions}           & \multicolumn{2}{c}{4}              &  &  & \multicolumn{2}{c}{5}       \\
\multicolumn{2}{c}{\#Operation notes}           & \multicolumn{2}{c}{13}             &  &  & \multicolumn{2}{c}{14}      \\ \hline
\end{tabular}
\end{table}

\subsubsection{PitVQA-Net:}
\begin{figure}[t]
    \centering
    \includegraphics[width=1\textwidth]{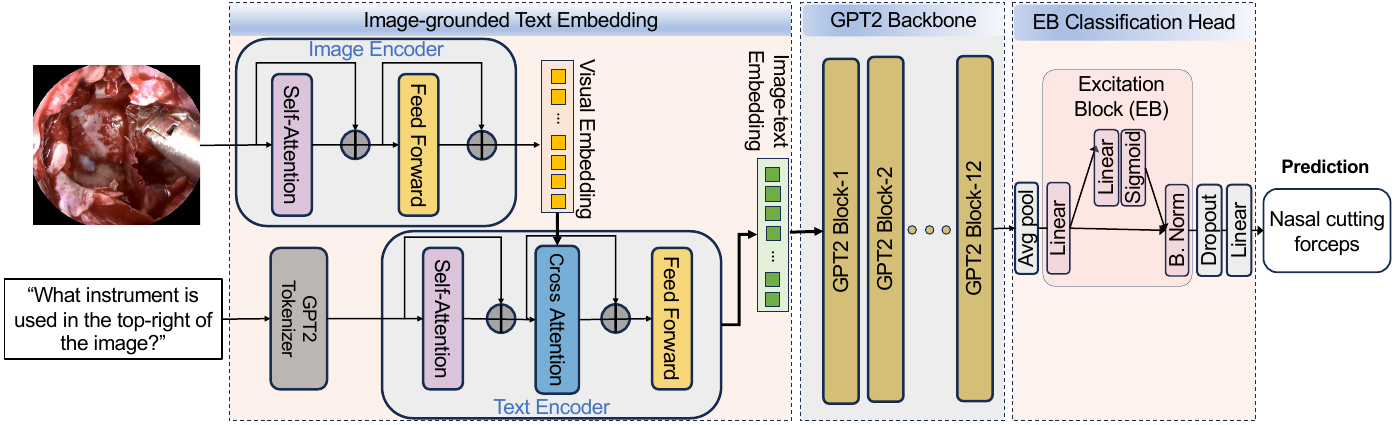}
    \caption{PitVQA-Net: The network forms of Image-grounded Text Embedding, GPT2 Backbone and Classification Head. The image-grounded text embedding leverages joint embedding, cross-attention and contextual representation.}
    \label{fig:model_archi}
\end{figure}

Our PitVQA-Net comprises an Image-grounded Text Embedding, a GPT2 Backbone, and an Excitation Block(EB) Classification Head as shown in the Fig.~\ref{fig:model_archi}. Detailed descriptions of these modules can be found below:\\

\noindent \textit{Image-grounded Text Embedding:}
Image-grounded Text Embedding forms of image encoder and cross-attention text encoder. The image encoder is a vision transformer which analyzes visual content and produces a detailed representation of the image features. The cross-attention text encoder (e.g. Bert Model~\cite{devlin2018bert}) processes the text input with the interaction of the image features through cross-attention. During the process, the image features attend to relevant parts of the text sequence, and textual features attend to specific image regions. This helps the model "ground" the language in corresponding visual elements. Through multiple layers of cross-attention, the image-grounded text encoder produces updated representations where both the text and image features are deeply contextualized with respect to one another. The output of the Image-grounded Text Encoder is a set of text embeddings that have been enriched with visual context. In our PiyVQA-Net, these embeddings then pass to the GPT2, followed by the EB classification head for the VQA prediction.\\

\noindent \textit{GPT2 Backbone:} 
The GPT2 backbone consists of 12 GPT2 blocks, which are the decoder-only transformer blocks. It is specifically designed for language generation tasks. Each GPT2 block contains multi-head attention, followed by a layer normalization layer and a feed-forward neural network. Multi-head attention computes self-attention multiple times in parallel with different learned projections of the query (Q), key (K), and value (V) vectors, then concatenates and linearly projects the results to produce a final representation. The self-attention~\cite{vaswani2017attention} mechanism is formulated as $\text{Attention}(Q, K, V) = \text{softmax}\left(\frac{QK^T}{\sqrt{d_k}}\right)V$.
It effectively captures and processes different aspects of the input data, leveraging the self-attention mechanism to model the dependencies between tokens in the sequence. The stacking of multiple such blocks allows GPT2 to model complex language patterns and generate coherent and contextually relevant text based on the input provided.
In our PitVQA-Net, the GPT2 backbone receives the embedding from our Image-grounded Text Embedding module, then processes them and produces hidden state features. These features are then passed to the SE classification head to predict the answer class in our VQA task.\\

\noindent \textit{EB Classification Head:} 
We design a lightweight Excitation Block (EB) layer with a linear layer, gated attention, and a batch normalization layer. The gated-attention forms of a sigmoid activation succeed a linear layer, which is capable of amplifying the significant features and suppressing the weak features by multiplying feature maps with gated weights obtained from the sigmoid function. We integrate EB into our classification head, including an average pooling, dropout, and a linear layer, as illustrated in Fig.~\ref{fig:model_archi}. In our PitVQA-Net, the feature maps from GPT2 Backbone are passed through the EB classification head to transform the feature maps into logits. Then, we use a softmax function to obtain the final probability distribution.

\section{Experiments and Results}

\subsection{Dataset}
In addition to our PitVQA dataset (details in the section~\ref{pitvqa_dataset}), we also validate our model on a publicly available dataset of EndoVis18-VQA~\cite{seenivasan2023surgicalgpt}. The dataset consists of 11,783 Q\&A pairs derived from 2,007 surgical scenes from 14 video sequences of nephrectomy surgery procedures. The answers are in the form of single words with 18 distinct labels (1 kidney, 13 tool-tissue interactions, and 4 instrument locations). We split the training and validation set by following the original setup~\cite{seenivasan2023surgicalgpt}. Thus, the training set contains 1560 frames and 9014 question-answer pairs, while the validation set consists of 447 frames and 2769 question-answer pairs. Both PitVQA and EndoVis18-VQA exhibit significant class imbalance, which limits the reliability of traditional accuracy metrics for robustness evaluation across the classes.

\subsection{Implementation Details}
The implementation and pre-trained weights of our backbone networks are adopted from the official repositories of the BLIP~\cite{li2022blip} and Huggingface GPT2\footnote{https://huggingface.co/openai-community/gpt2}. Our model is trained on cross-entropy loss and Adam optimizer with the learning rate of $1 \times 10^{-5}$. For the performance comparison, we selected closely related state-of-the-art (SOTA) surgical VQA models such as SurgicalGPT~\cite{seenivasan2023surgicalgpt}, VisualBert RM~\cite{seenivasan2022surgical}, and VisualBert~\cite{li2019visualbert} to retrain using official repositories. Additionally, we adopted the results of other recent baselines, including MFH~\cite{yu2018beyond}, MFB~\cite{yu2017multi}, and Mutan~\cite{ben2017mutan}, as reported in~\cite{seenivasan2023surgicalgpt}. All experiments are conducted with the PyTorch framework on an NVIDIA RTX A6000 GPU.

\subsection{Evaluation Metrics}
To demonstrate the model's generalizability and robustness across datasets with imbalanced class frequency, balanced accuracy has proven to be an effective evaluation metric~\cite{maier2024metrics}. It is a prevalence-independent measure that computes the prediction accuracy by equally weighting the contribution of each class. Given the significant class imbalance in the datasets, we have employed metrics such as balanced accuracy (B. Acc), and Fscore, including Accuracy and Recall.

\subsection{Results}

Table~\ref{table:comparison} presents the results of the proposed method compared with other SOTA surgical VQA models. We emphasize on the metric of balanced accuracy to highlight the robustness of the model prediction across the classes considering highly imbalanced dataset. There are significant performance improvements in our method with a balanced accuracy of 58.82\% and 44.46\% for the datasets of PitVQA and EndoVis18-VQA, respectively. The performance improvements are around 8\% on PitVQA and 9\% on EndoVis18-VQA over the closely related work of SurgicalGPT. Similar trends are observed in other metrics of Fscore and recall.

\begin{figure}[!t]
    \centering
    \includegraphics[width=1\textwidth]{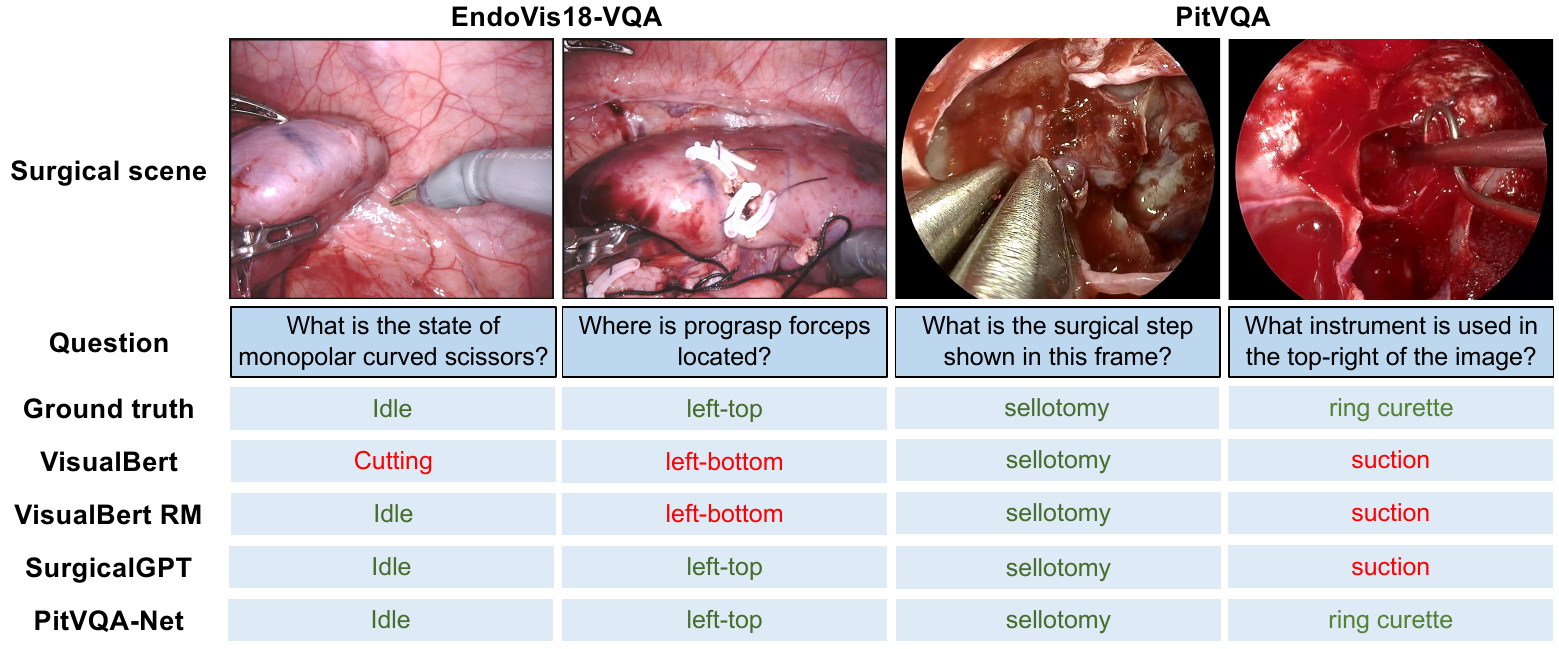}
    \caption{Qualitative visualization of our model prediction comparing with closely related works with datasets of our PitVQA and EndoVis18-VQA.}
    \label{fig:qualitative}
\end{figure}

The qualitative performance comparison for both PitVQA and EndoVis18-VQA is illustrated in Fig.~\ref{fig:qualitative}. It appears that most of the models are capable of accurately recognizing surgical steps, whereas the identification of instruments with localization reasoning mostly fails. However, PitVQA-Net demonstrates robust prediction across various types of question answering in both datasets.

\begin{table}[]
\centering
\caption{The performance comparison of our method with other SOTA surgical VQA models in our PitVQA dataset. The balanced accuracy~\cite{maier2024metrics} is denoted as B. Acc.}
\label{table:comparison}
\scalebox{0.94}{
\begin{tabular}{c|cccc|cccc}
\hline
\multirow{2}{*}{\textbf{MODELS}} &
  \multicolumn{4}{c|}{\textbf{EndoVis18-VQA}} &
  \multicolumn{4}{c}{\textbf{PitVQA}} \\ \cline{2-9} 
 &
  \multicolumn{1}{c|}{\textbf{FScore}} &
  \multicolumn{1}{c|}{\textbf{B. Acc}} &
  \multicolumn{1}{c|}{\textbf{Acc}} &
  \textbf{Recall} &
  \multicolumn{1}{c|}{\textbf{FScore}} &
  \multicolumn{1}{c|}{\textbf{B. Acc}} &
  \multicolumn{1}{c|}{\textbf{Acc}} &
  \textbf{Recall} \\ \hline
\textbf{Mutan~\cite{ben2017mutan}} &
  \multicolumn{1}{c|}{0.4565} &
  \multicolumn{1}{c|}{---} &
  \multicolumn{1}{c|}{0.6303} &
  \textbf{0.4969} &
  \multicolumn{1}{c|}{---} &
  \multicolumn{1}{c|}{---} &
  \multicolumn{1}{c|}{---} &
  --- \\
\textbf{MFB~\cite{yu2017multi}} &
  \multicolumn{1}{c|}{0.3622} &
  \multicolumn{1}{c|}{---} &
  \multicolumn{1}{c|}{0.5238} &
  0.4205 &
  \multicolumn{1}{c|}{---} &
  \multicolumn{1}{c|}{---} &
  \multicolumn{1}{c|}{---} &
  --- \\
\textbf{MFH~\cite{yu2018beyond}} &
  \multicolumn{1}{c|}{0.4224} &
  \multicolumn{1}{c|}{---} &
  \multicolumn{1}{c|}{0.5876} &
  0.4835 &
  \multicolumn{1}{c|}{---} &
  \multicolumn{1}{c|}{---} &
  \multicolumn{1}{c|}{---} &
  --- \\
\textbf{VisualBert~\cite{li2019visualbert}} &
  \multicolumn{1}{c|}{0.3745} &
  \multicolumn{1}{c|}{0.3474} &
  \multicolumn{1}{c|}{0.6143} &
  0.4282 &
  \multicolumn{1}{c|}{0.4286} &
  \multicolumn{1}{c|}{0.4358} &
  \multicolumn{1}{c|}{0.6338} &
  0.4549 \\
\textbf{VisualBert RM~\cite{seenivasan2022surgical}} &
  \multicolumn{1}{c|}{0.3583} &
  \multicolumn{1}{c|}{0.3422} &
  \multicolumn{1}{c|}{0.6190} &
  0.4079 &
  \multicolumn{1}{c|}{0.4281} &
  \multicolumn{1}{c|}{0.3892} &
  \multicolumn{1}{c|}{0.6318} &
  0.4103 \\
\textbf{SurgicalGPT~\cite{seenivasan2023surgicalgpt}} &
  \multicolumn{1}{c|}{0.4649} &
  \multicolumn{1}{c|}{0.3543} &
  \multicolumn{1}{c|}{0.6811} &
  0.4649 &
  \multicolumn{1}{c|}{0.5261} &
  \multicolumn{1}{c|}{0.5090} &
  \multicolumn{1}{c|}{0.7232} &
  0.5397 \\ \hline
\textbf{PitVQA-Net (Ours)} &
  \multicolumn{1}{c|}{\textbf{0.6204}} &
  \multicolumn{1}{c|}{\textbf{0.4446}} &
  \multicolumn{1}{c|}{\textbf{0.6829}} &
  0.4793 &
  \multicolumn{1}{c|}{\textbf{0.5952}} &
  \multicolumn{1}{c|}{\textbf{0.5882}} &
  \multicolumn{1}{c|}{\textbf{0.7601}} &
  \textbf{0.5917} \\ \hline
\end{tabular}
}
\end{table}

\subsection{Ablation Study}
To assess the effectiveness of our proposed method, we conducted an ablation study comparing different vision-language embeddings, including CLIP~\cite{radford2021learning}, as well as other large language model variants such as BioGPT~\cite{luo2022biogpt} and BERT~\cite{devlin2018bert}, as detailed in Table~\ref{table:ablation_study}. Additionally, we explored the benefits of using LLM pre-trained weights by conducting an experiment with PitVQA-Net without leveraging any pre-trained weights. We also investigate the effect of EB in our PitVQA-Net. Our experiments show that our EB block significantly enhances performance on the EndoVis18-VQA dataset. Overall, the results demonstrate the effectiveness of each innovation within our proposed method.

\begin{table}[!h]
\centering
\caption{Ablation study with other vision-language embedding of CLIP~\cite{radford2021learning} and LLMs like BioGPT~\cite{luo2022biogpt}, and BERTZ\cite{devlin2018bert}. We also observe the performance of the pitVQA-Net without pretrained weights (w/o pretrain.), and without excitation block (w/o EB).}
\label{table:ablation_study}
\begin{tabular}{cll|ccc|ccc}
\hline
\multicolumn{3}{c|}{\multirow{2}{*}{MODELS}} &
  \multicolumn{3}{c|}{EndoVis18-VQA} &
  \multicolumn{3}{c}{PitVQA} \\ \cline{4-9} 
\multicolumn{3}{c|}{} &
  \multicolumn{1}{c|}{B. Acc} &
  \multicolumn{1}{c|}{Recall} &
  FScore &
  \multicolumn{1}{c|}{B. Acc} &
  \multicolumn{1}{c|}{Recall} &
  FScore \\ \hline
\multicolumn{3}{c|}{CLIP-GPT} &
  \multicolumn{1}{c|}{0.4342} &
  \multicolumn{1}{c|}{\textbf{0.5503}} &
  0.4513 &
  \multicolumn{1}{c|}{0.4138} &
  \multicolumn{1}{c|}{0.4405} &
  0.4176 \\
\multicolumn{3}{c|}{CLIP-BioGPT} &
  \multicolumn{1}{c|}{0.4145} &
  \multicolumn{1}{c|}{{0.4876}} &
  0.4728 &
  \multicolumn{1}{c|}{0.4072} &
  \multicolumn{1}{c|}{0.4295} &
  0.4168 \\
\multicolumn{3}{c|}{BLIP-BERT} &
  \multicolumn{1}{c|}{0.3752} &
  \multicolumn{1}{c|}{0.4525} &
  {0.5317} &
  \multicolumn{1}{c|}{{0.5574}} &
  \multicolumn{1}{c|}{0.5957} &
  {0.5663} \\ \hline
\multicolumn{3}{c|}{PitVQA-Net (w/o pretrain.)} &
  \multicolumn{1}{c|}{{0.3650}} &
  \multicolumn{1}{c|}{0.4265} &
  0.5345 &
  \multicolumn{1}{c|}{0.5419} &
  \multicolumn{1}{c|}{0.5710} &
  0.5647 \\
\multicolumn{3}{c|}{PitVQA-Net (w/o EB)} &
  \multicolumn{1}{c|}{{0.3978}} &
  \multicolumn{1}{c|}{0.4577} &
  0.5523 &
  \multicolumn{1}{c|}{0.5800} &
  \multicolumn{1}{c|}{{\textbf{0.5982}}} &
  0.5939 \\ 
\multicolumn{3}{c|}{PitVQA-Net} &
  \multicolumn{1}{c|}{{\textbf{0.4446}}} &
  \multicolumn{1}{c|}{0.4793} &
  \textbf{0.6204} &
  \multicolumn{1}{c|}{\textbf{0.5882}} &
  \multicolumn{1}{c|}{{0.5917}} &
  \textbf{0.5952} \\ \hline
\end{tabular}
\end{table}

\section{Discussion and Conclusion}
In this study, we introduced PitVQA, a targeted dataset for VQA in the domain of endonasal pituitary adenoma surgery and PitVQA-Net, an innovative adaptation of the GPT2 that incorporates a novel image-grounded text embedding and gated-attention excitation block. Our experimental results demonstrate a clear performance advantage for PitVQA-Net, achieving superior results on both the PitVQA dataset and the publicly available EndoVis18-VQA dataset when compared to existing surgical VQA models. The ablation study further reinforces the significance of our proposed method, highlighting the effectiveness of our image-grounded text embedding, excitation block, selection of LLM, and importance of pretrained weights. This work paves the way for the development of intuitive and collaborative surgical AI assistants. By enabling accurate and contextually aware responses to complex surgical questions, PitVQA-Net demonstrates significant promise for enhancing intra-operative decision-making and ultimately improving patient outcomes. Future directions for this research include expanding the PitVQA dataset to generate sentences by covering a wider range of surgical scenarios, exploring more advanced vision-language fusion techniques, and investigating the potential for real-time deployment of such systems within the operating room.

\subsection*{Acknowledgements}
This work was supported in whole, or in part, by the Wellcome/EPSRC Centre for Interventional and Surgical Sciences (WEISS) [203145/Z/16/Z] and the Engineering and Physical Sciences Research Council (EPSRC) [EP/W00805X/1, EP/Y01958X/1]; Horizon 2020 FET Open [863146]; the UCLH/UCL NIHR Biomedical Research Centre; the Department of Science, Innovation and Technology (DSIT); and the Royal Academy of Engineering Chair in Emerging Technologies Scheme. AD is supported by EPSRC [EP/S021612/1]. With thanks to Digital Surgery Ltd, a Medtronic company, for access to Touch Surgery\textsuperscript{TM} Enterprise for both video recording and storage.

\bibliography{mybib}{}

\begin{thebibliography}{10}
\providecommand{\url}[1]{\texttt{#1}}
\providecommand{\urlprefix}{URL }
\providecommand{\doi}[1]{https://doi.org/#1}

\bibitem{antol2015vqa}
Antol, S., Agrawal, A., Lu, J., Mitchell, M., Batra, D., Zitnick, C.L., Parikh, D.: Vqa: Visual question answering. In: Proceedings of the IEEE international conference on computer vision. pp. 2425--2433 (2015)

\bibitem{bai2023cat}
Bai, L., Islam, M., Ren, H.: Cat-vil: Co-attention gated vision-language embedding for visual question localized-answering in robotic surgery. In: International Conference on Medical Image Computing and Computer-Assisted Intervention. pp. 397--407. Springer (2023)

\bibitem{bai2023surgicalvqla}
Bai, L., Islam, M., Seenivasan, L., Ren, H.: Surgical-vqla: Transformer with gated vision-language embedding for visual question localized-answering in robotic surgery (2023)

\bibitem{ben2017mutan}
Ben-Younes, H., Cadene, R., Cord, M., Thome, N.: Mutan: Multimodal tucker fusion for visual question answering. In: Proceedings of the IEEE international conference on computer vision. pp. 2612--2620 (2017)

\bibitem{das2023multi}
Das, A., Khan, D.Z., Williams, S.C., Hanrahan, J.G., Borg, A., Dorward, N.L., Bano, S., Marcus, H.J., Stoyanov, D.: A multi-task network for anatomy identification in endoscopic pituitary surgery. In: International Conference on Medical Image Computing and Computer-Assisted Intervention. pp. 472--482. Springer (2023)

\bibitem{decker2023large}
Decker, H., Trang, K., Ramirez, J., Colley, A., Pierce, L., Coleman, M., Bongiovanni, T., Melton, G.B., Wick, E.: Large language model- based chatbot vs surgeon-generated informed consent documentation for common procedures. JAMA Network Open  \textbf{6}(10),  e2336997--e2336997 (2023)

\bibitem{devlin2018bert}
Devlin, J., Chang, M.W., Lee, K., Toutanova, K.: Bert: Pre-training of deep bidirectional transformers for language understanding. arXiv preprint arXiv:1810.04805  (2018)

\bibitem{hu2018squeeze}
Hu, J., Shen, L., Sun, G.: Squeeze-and-excitation networks. In: Proceedings of the IEEE conference on computer vision and pattern recognition. pp. 7132--7141 (2018)

\bibitem{hudson2019gqa}
Hudson, D.A., Manning, C.D.: Gqa: A new dataset for real-world visual reasoning and compositional question answering. In: Proceedings of the IEEE/CVF conference on computer vision and pattern recognition. pp. 6700--6709 (2019)

\bibitem{khan2023current}
Khan, D.Z., Hanrahan, J.G., Baldeweg, S.E., Dorward, N.L., Stoyanov, D., Marcus, H.J.: Current and future advances in surgical therapy for pituitary adenoma. Endocrine Reviews  (2023)

\bibitem{lawson2023artificial}
Lawson~McLean, A.: Artificial intelligence in surgical documentation: A critical review of the role of large language models. Annals of Biomedical Engineering pp.~1--2 (2023)

\bibitem{li2022blip}
Li, J., Li, D., Xiong, C., Hoi, S.: Blip: Bootstrapping language-image pre-training for unified vision-language understanding and generation. In: International Conference on Machine Learning. pp. 12888--12900. PMLR (2022)

\bibitem{li2019visualbert}
Li, L.H., Yatskar, M., Yin, D., Hsieh, C.J., Chang, K.W.: Visualbert: A simple and performant baseline for vision and language. arXiv preprint arXiv:1908.03557  (2019)

\bibitem{luo2022biogpt}
Luo, R., Sun, L., Xia, Y., Qin, T., Zhang, S., Poon, H., Liu, T.Y.: Biogpt: generative pre-trained transformer for biomedical text generation and mining. Briefings in Bioinformatics  \textbf{23}(6),  bbac409 (2022)

\bibitem{maier2024metrics}
Maier-Hein, L., Reinke, A., Godau, P., Tizabi, M.D., Buettner, F., Christodoulou, E., Glocker, B., Isensee, F., Kleesiek, J., Kozubek, M., et~al.: Metrics reloaded: recommendations for image analysis validation. Nature methods pp. 1--18 (2024)

\bibitem{marcus2021pituitary}
Marcus, H.J., Khan, D.Z., Borg, A., Buchfelder, M., Cetas, J.S., Collins, J.W., Dorward, N.L., Fleseriu, M., Gurnell, M., Javadpour, M., et~al.: Pituitary society expert delphi consensus: operative workflow in endoscopic transsphenoidal pituitary adenoma resection. Pituitary  \textbf{24}(6),  839--853 (2021)

\bibitem{radford2021learning}
Radford, A., Kim, J.W., Hallacy, C., Ramesh, A., Goh, G., Agarwal, S., Sastry, G., Askell, A., Mishkin, P., Clark, J., et~al.: Learning transferable visual models from natural language supervision. In: International conference on machine learning. pp. 8748--8763. PMLR (2021)

\bibitem{radford2019language}
Radford, A., Wu, J., Child, R., Luan, D., Amodei, D., Sutskever, I., et~al.: Language models are unsupervised multitask learners. OpenAI blog  \textbf{1}(8), ~9 (2019)

\bibitem{roy2018concurrent}
Roy, A.G., Navab, N., Wachinger, C.: Concurrent spatial and channel ‘squeeze \& excitation’in fully convolutional networks. In: Medical Image Computing and Computer Assisted Intervention--MICCAI 2018: 21st International Conference, Granada, Spain, September 16-20, 2018, Proceedings, Part I. pp. 421--429. Springer (2018)

\bibitem{seenivasan2023surgicalgpt}
Seenivasan, L., Islam, M., Kannan, G., Ren, H.: Surgicalgpt: End-to-end language-vision gpt for visual question answering in surgery. arXiv preprint arXiv:2304.09974  (2023)

\bibitem{seenivasan2022surgical}
Seenivasan, L., Islam, M., Krishna, A.K., Ren, H.: Surgical-vqa: Visual question answering in surgical scenes using transformer. In: International Conference on Medical Image Computing and Computer-Assisted Intervention. pp. 33--43. Springer (2022)

\bibitem{vaswani2017attention}
Vaswani, A., Shazeer, N., Parmar, N., Uszkoreit, J., Jones, L., Gomez, A.N., Kaiser, {\L}., Polosukhin, I.: Attention is all you need. Advances in neural information processing systems  \textbf{30} (2017)

\bibitem{yu2017multi}
Yu, Z., Yu, J., Fan, J., Tao, D.: Multi-modal factorized bilinear pooling with co-attention learning for visual question answering. In: Proceedings of the IEEE international conference on computer vision. pp. 1821--1830 (2017)

\bibitem{yu2018beyond}
Yu, Z., Yu, J., Xiang, C., Fan, J., Tao, D.: Beyond bilinear: Generalized multimodal factorized high-order pooling for visual question answering. IEEE transactions on neural networks and learning systems  \textbf{29}(12),  5947--5959 (2018)

\bibitem{yuan2024advancing}
Yuan, K., Kattel, M., Lavanchy, J.L., Navab, N., Srivastav, V., Padoy, N.: Advancing surgical vqa with scene graph knowledge (2024)

\end{thebibliography}
\bibliographystyle{splncs04}

\end{document}